\documentclass[conference]{IEEEtran}
\IEEEoverridecommandlockouts
\usepackage{cite}
\usepackage{amsmath,amssymb,amsfonts}
\usepackage{algorithmic}
\usepackage{graphicx}
\usepackage{textcomp}
\usepackage{xcolor}

\usepackage{verbatim}
\usepackage{latexsym}
\usepackage{threeparttable}
\usepackage{algorithmic}
\usepackage[linesnumbered,commentsnumbered,ruled,vlined]{algorithm2e}
\usepackage{adjustbox}
\usepackage{multicol}
\usepackage{amsmath}
\usepackage{setspace}
\usepackage{comment}
\usepackage{float}
\usepackage{multirow}
\usepackage{xparse}
\usepackage{xcolor,colortbl}
\usepackage{booktabs}
\usepackage{tabularx} 
\usepackage{enumitem}
\newlist{steps}{enumerate}{1}
\setlist[steps, 1]{label = Step \arabic*:}
\usepackage{booktabs}
\renewcommand{\arraystretch}{1.3}
\usepackage{longtable}
\usepackage{translator}
\floatstyle{plaintop}
\usepackage{subcaption}
\usepackage[english]{babel}
\usepackage[autostyle]{csquotes}
\MakeOuterQuote{"}

\def\BibTeX{{\rm B\kern-.05em{\sc i\kern-.025em b}\kern-.08em
    T\kern-.1667em\lower.7ex\hbox{E}\kern-.125emX}}
\begin{document}

\title{Synthetic Data Generation for Fraud Detection using GANs\\
{\footnotesize \textsuperscript{*}}
}

\author{\IEEEauthorblockN{Charitos Charitou}
\IEEEauthorblockA{\textit{Department of Computer Science} \\
\textit{City, University of London}\\
London, UK \\
charitos.charitou@city.ac.uk }
\and
\IEEEauthorblockN{ Simo Dragicevic}
\IEEEauthorblockA{\textit{BetBuddy} \\
\textit{Playtech Plc}\\
London, UK \\
simo.dragicevic@playtech.com}
\and
\IEEEauthorblockN{Artur d'Avila Garcez}
\IEEEauthorblockA{\textit{Department of Computer Science} \\
\textit{City, University of London}\\
London, UK \\
a.garcez@city.ac.uk}

}

\maketitle

\begin{abstract}

Detecting money laundering in gambling is becoming increasingly challenging for the gambling industry as consumers migrate to online channels. Whilst increasingly stringent regulations have been applied over the years to prevent money laundering in gambling, despite this, online gambling is still a channel for criminals to spend proceeds from crime. Complementing online gambling's growth more concerns are raised to its effects compared with gambling in traditional, physical formats, as it might introduce higher levels of problem gambling or fraudulent behaviour due to its nature of immediate interaction with online gambling experience. However, in most cases the main issue when organisations try to tackle those areas is the absence of high quality data. Since fraud detection related issues face the significant problem of the class imbalance, in this paper we propose a novel system based on Generative Adversarial Networks (GANs) for generating synthetic data in order to train a supervised classifier. Our framework Synthetic Data Generation GAN (SDG-GAN), manages to outperformed density based over-sampling methods and improve the classification performance of benchmarks datasets and the real world gambling fraud dataset.

\end{abstract}


\section{Introduction}

The major motivations for this research are to understand the problems faced by the gambling industry with regard to raising standards in money laundering detection. The AML process is conceptually similar to fraud detection, an area that has been the focus of a great deal of research in recent years. It has been shown that applying machine learning techniques to detect fraud can solve the problem to a certain degree, with the best results achieved with supervised learning. However, the problem with supervised learning is that it requires labelled data for both non-fraudulent and fraudulent behaviours in order to train a model \cite{charitou2020semi}. Collaborating with online gambling operator which supported this research by making anonymous data available with labels describing high-risk (fraudulent customers) and no-risk of money laundering (non-fraudulent customers). Notwithstanding, the non-fraudulent customers are much greater in number compared to customers with high money laundering risk.

Most supervised learning algorithms are not designed to cope with a large difference in the number of cases belonging to different classes \cite{batista2000applying}. This problem is known in the literature as class imbalance and is an issue regularly encountered by researchers. The problem corresponds to the issue faced by inductive learning systems when dealing with domains where one class is represented by a large number of samples while the other class is represented by fewer samples. In such cases, the reliability and validity of the results are questionable since prediction algorithms tend to have a bias towards the majority class.  Such an imbalanced dataset could lead to unintended model performance – for example, classifying all the cases as normal and managing to achieve almost perfect accuracy. This, however, is not helpful in real-world situations. Therefore, the problem arises of how to improve the identification of the minority class as opposed to achieving higher overall accuracy. The class imbalance problem has been the subject of extensive research \cite{chawla2009data},\cite{ chawla2004special},\cite{batuwita2013class} in different areas i.e. fraud detection, healthcare.

Many techniques for handling imbalanced data have emerged in the literature \cite{kubat1997addressing}, \cite{SMOTE},\cite{adasynCite},\cite{han2005borderline}.  Solutions have been implemented at the algorithmic, data and hybrid level. At the algorithmic level, algorithms are adjusted in order to reduce bias towards the majority class and improve classification. At the data level, sampling techniques are applied for synthetic data generation to balance the dataset. Finally, hybrid-level approaches combine data-level and algorithmic-level techniques. In Section \ref{literature}, we present the relevant literature. The class imbalance issue is observed in binary and multi-class classification problems. In this paper we focus on the binary classification problem as we try tackle the issue of fraud in the online gambling space.

A direct approach to the data generation process would be the use of a generative model that captures the actual data distribution \cite{douzas2018effective} for generating synthetic data. Generative adversarial networks (GAN) are a recent method that uses neural networks to create generative models \cite{GANs}. As previous studies have shown, GANs can be used effectively as an oversampling method to produce high-quality synthetic data \cite{tanaka2019data}. In contrast with other generative techniques, GANs are able to parallelise sample generation with sample classification. Further, they make no assumption about distribution and variational bounds. Finally, GANs make no use of Markov chain or maximum likelihood estimation \cite{GANs,ngwenduna2021alleviating}.

In this paper, we propose a GAN-based approach called synthetic data generation GAN (SDG-GAN), which, as the empirical results show, can be a powerful tool for tackling the imbalanced class problem on structured data by generating new high-quality instances. In Section \ref{GANINTRO} an overview of GANs is provided, where in Section \ref{methodExplain} we introduce our approach. Our method is validated in Section \ref{methodValid} and \ref{Results_oversa} via experiments on benchmark datasets (Credit Card Fraud, Breast Cancer Wisconsin, Pima Diabetes) from different disciplines before applying it for generating new synthetic data for online gambling players in Section \ref{sdgog}. Our method is evaluated in terms of its classification performance when combined with the classification models, namely logistic regression (LR), random forest (RF), multi-layer Perceptron (MLP) and XGBoost (XGB). 

\section{Related Work} \label{literature}

Data-level methods are described as the sampling techniques used to balance a dataset \cite{alhakbani2019handling}. This means that the number of instances of each class is adjusted either by increasing the instances of the minority class or by decreasing the instances of the majority class. In general, applying sampling algorithms will result in the alteration of the distribution of an imbalanced dataset until it is balanced. Various studies have shown that a balanced dataset can improve the performance of a classifier \cite{japkowicz2011evaluating,garcia2015data}. Oversampling, undersampling and hybrid methods have been applied to achieve a balanced dataset. 

 Synthetic oversampling (i.e. generating new synthetic instances) and random oversampling (ROS) are the two methods of oversampling. In ROS, minority samples are added to the training set by randomly replicating minority class samples. Although the performance of a prediction algorithm can be improved with ROS \cite{IC5}, Chawla \cite{IC2}  has suggested that it could also cause overfitting – since the same data may be used more than once – and could be more computationally expensive. 

Notwithstanding the problems originating from ROS, advancements in the field of imbalanced classification show that most issues can be overcome with synthetic oversampling. Synthetic oversampling methods generate new synthetic instances in order to balance a dataset. Examples of synthetic oversampling techniques include but are not limited to ADAptive SYNthetic sampling (ADASYN) \cite{adasynCite} and Synthetic Minority Oversampling TEchnique (SMOTE) \cite{SMOTE}. A popular extension to SMOTE includes selecting instances of the minority class that are misclassified, such as with a k-nearest neighbour classification model. This modified SMOTE method is called Bordeline-SMOTE (B-SMOTE) \cite{han2005borderline}.

The key difference between ADASYN and SMOTE is that ADASYN uses a density distribution criterion to automatically decide how many samples need to be generated for each minority data point. First, it improves learning by reducing the bias caused by the imbalance in class priors. Second, it improves performance because the classification decision boundary is adaptively shifted toward ‘difficult examples’\cite{adasynCite}.

Finally, hybrid methods incorporate both oversampling and undersampling techniques. Ganguly and Sadaoui \cite{ganguly2017classification} utilised a hybrid method of data oversampling and undersampling to improve effectiveness in addressing the issue of highly imbalanced auction fraud datasets. Their results showed a significant classification improvement for various well-known classifiers. Other popular hybrid methods in the literature include SMOTE$+$TOMEK  \cite{batista2003balancing} and  SMOTE$+$ENN \cite{batista2004study}. 
SMOTE$+$TOMEK  aims to clean overlapping data points for each of the classes distributed in sample space, while SMOTE$+$ ENN deletes any  instance of the majority class which its nearest neighbours are misclassified.

GANs are one of the most popular and successful generative technique for synthetic data generation \cite{GANs}, especially image generation \cite{zhai2019lifelong}\cite{xu2018attngan}.  Literature on using GANs for oversampling structured data has also emerged. Douzas and Bacao \cite{douzas2018effective} used a conditional GAN (cGAN) to approximate the true data distribution and generate data for the minority classes of various imbalanced datasets. They compared their results against standard oversampling approaches and showed improvements in the quality of data generation. 

Lei et al. \cite{xu2019modeling} designed CTGAN, a cGAN-based method to balance tabular datasets with both continuous and discrete columns. They designed a benchmark with seven simulated and eight real datasets and several Bayesian network baselines. CTGAN outperformed Bayesian methods on most of the real datasets while other deep learning methods did not. The authors in \cite{fiore2019using} proposed oversampling by training a GAN with vanilla GAN loss on only minority class observations. They compared their method against SMOTE and no oversampling and reported mixed results. Experiments showed that a classifier trained on the augmented dataset outperformed the same classifier trained on the original data.  In this work, GANs are examined  for tackling the imbalanced class issue, through the generation of synthetic data. 

\section{GANs} \label{GANINTRO}

GANs are generative models based on a game-theoretic scenario in which a generator $(G)$ network is competing against a discriminator $(D)$ \cite{GANs}. The generator, with noise variable $Z$ as input, generates fake samples with distribution $p_{g}$ that match the true data distribution, $p_{data}$. However, the discriminator network is trained to distinguish the real samples (drawn from the training data) and fake samples generated from G \cite{charitou2020semi}.

A common analogy in the literature for GANs \cite{creswell2018generative} is to think of one network as an art forger and another as an art expert. The forger, known in the literature as the generator, G, creates forgeries with the aim of making realistic images. The expert, known as the discriminator, D, receives both forgeries and real images and aims to tell them apart. Both are trained simultaneously and in competition with each other.

Typically, the discriminator model is trained to maximise its ability to distinguish real input data from fake data. The generator tries to fool the discriminator by producing better fake samples. Mathematically, the generator and discriminator play a min-max two-player game with value function V(G,D) \cite{GANs}:
\begin{equation} \label{gan}
\begin{aligned}
\underset{G}{min} \ \underset{D}{max} \ V(G,D) = E_{x\sim p_{data}}[log(D(x))] \\ + E_{z \sim p_{z}}[1-log(D(G(z)))]  \end{aligned}
\end{equation}
\noindent where E is the expectation, $p_{data}$ is the real data distribution and $p_{z}$ is a noise distribution. The training of a GAN could be characterised as an optimisation process for both  the generator and discriminator. The output of the generator is defined as $p_{g}$. As equation (\ref{gan}) suggests, GANs aim to minimise the Jensen–Shannon divergence between the data distribution $p_{data}$ and the generative distribution $p_{g}$ with perfect minimisation reached when $p_{g} = p_{data}$. The optimisation equations for the generator and the discriminator are defined respectively as follows:
\begin{equation} \label{generator}
\begin{aligned}
\underset{G}{min}\  E_{z \sim p_{z}}[1-log(D(G(z))) \end{aligned}
\end{equation}
\begin{equation} \label{discriminator_1}
\begin{aligned}
\underset{D}{max} \ E_{x\sim p_{data}}[log(D(x))] \\ + E_{z \sim p_{z}}[1-log(D(G(z)))]  \end{aligned}
\end{equation}

Although GANs are very promising for synthetic data generation, training a GAN is challenging \cite{GANs} and often unstable which could lead to the following `symptoms' \cite{radford2015unsupervised,b4,li2017towards}:

\begin{itemize}
\item Difficulties in making both the discriminator and generator converge \cite{radford2015unsupervised}.
\item Collapse of the generator model by producing similar samples from different inputs \cite{b4}.
\item The discriminator converging quickly to zero \cite{li2017towards}, providing no reliable path for gradient updates to the generator.
\end{itemize}

Researchers have considered several approaches to overcome such issues. They have been experimenting with architectural changes \cite{b4}, different loss functions \cite{arjovsky2017wasserstein} and both. Our SDG-GAN tries to eliminate the above problems by combining a different loss function and architecture compared to the original vanilla GAN \cite{GANs}. In Section \ref{conditional_GANs}, we give an overview of conditional GANs (cGANs) \cite{engelmann2020conditional} and in Section \ref{methodExplain}, we explain how they were used to build the SDG-GAN architecture.

\subsection{Conditional GANs}\label{conditional_GANs}

Conditional GANs \cite{mirza2014conditional} are a simple extension of the original GAN framework, which conditions the generator on class labels to generate output for a specific class \cite{engelmann2020conditional}. The conditioning is achieved by feeding the class label $y$ into both the discriminator and generator as additional input. Thus, the generator estimates the distribution of $p_{X|y}$, and the discriminator learns to estimate $D(X,y) = P(fake|X,y)$. The modification of the generator and discriminator with the conditional rule allows for the generation of samples belonging to a specific class. Furthermore, the conditional discriminator ensures that the generator does not ignore class labels \cite{engelmann2020conditional}. Formally, the objective value function between generator G and discriminator D is the min-max in equation (\ref{cgan}).

\begin{equation} \label{cgan}
\begin{aligned}
\underset{G}{min} \ \underset{D}{max} \ V(G,D) = E_{x\sim p_{data}}[log(D(x|y))] \\ + E_{z \sim p_{z}}[1-log(D(G(z|y)))]  \end{aligned}
\end{equation}

During cGAN training, the discriminator is trained first with batches of only real features, $y_{real}=1$ and then with batches of only fake ones, $y_{fake}=0$ before the generator training continues through the GAN model. The GAN model assumes that the generated features will always be real, $y_{gan}=1$. As cGAN is an extension of the original generative adversarial framework, it exhibits the same problematic behavior, i.e. mode collapse and unstable training, due to the vanishing gradient problem \cite{zheng2020conditional}.

\section{Synthetic Data Generation GAN}\label{methodExplain}

In the  SDG-GAN framework  the generator and discriminator of SDG-GAN are both  feedforward networks with a MLP architecture.  The generator of a regular GAN aims to generate fake data  that are close to the real distribution. The discriminator of a regular GAN is used to identify whether an input is real  or  fake from the generator.


The process of generating new instances of the minority class requires training the GAN to estimate the distribution of the data. When the training phase is completed, new synthetic data can be generated utilising the generator’s abilities. The cGAN architecture of estimating the conditional distribution, $p_{x|y}$, is adapted in our method to generate the minority class samples. Instead of regular loss, feature matching loss is adapted by the SDG-GAN. Feature matching loss was introduced by \cite{b4} as a method for improving GAN training. 

Here, we propose a GAN architecture based on cGANs. The generator is a feedforward neural network that tries to learn the actual data distribution. In contrast with a cGAN generator, we use a feature matching technique to train the generator. Feature matching changes the cost function for the generator to minimise the statistical differences between the features of the real data and generated data. This changes the scope of the generative network from fooling the opponent to matching features in the real data. The objective function of feature matching loss is defined as follows:

\begin{equation} \label{lossFeature}
\begin{aligned}
||E_{x\sim p_{data}} f(x) - E_{z\sim p_{z}(z)}f(G(z))||^{2}_{2}
\end{aligned}
\end{equation}

where $f(x)$ is the feature vector extracted by an intermediate layer in the discriminator. Feature matching addresses the instability of GANs by specifying a new objective for the generator that prevents it from over-training. Instead of directly maximising the output of the discriminator, the new objective requires the generator to generate data that match the statistics of the real data, while we use the discriminator only to specify the statistics we think are worth matching. Specifically, we train the generator to match the expected value of the features on an intermediate layer of the discriminator. This is a natural choice of statistics for the generator to match because by training the discriminator, we ask it to find the features that are most discriminative of real data versus data generated by the current model \cite{b4}.

To oversample an imbalanced dataset, we first trained the SDG-GAN’s generator with imbalanced samples to estimate the data distribution. Once the training was completed, we could oversample the data by specifying to the generator how many new synthetic instances of the minority class we wanted to produce. We used a cGAN structure to estimate the conditional distribution, $p_{X|y}$, which allowed us to sample the minority class explicitly by conditioning the generator on the minority class label, $X_{new} = G(z, y = y^{minority})$.

The discriminator was trained similarly to a regular GAN discriminator. As with regular cGAN training, the objective had a fixed point where G exactly matched the distribution of the training data. We had no guarantee of reaching this fixed point in practice, but our empirical results indicated that feature matching is indeed effective in situations wherein a regular GAN becomes unstable \cite{b4}. Thus, we achieve the following objective function:

\begin{equation}\label{objectivefinal}
\begin{aligned}
\underset{G}{min} \ \underset{D}{max} \ \underset{FM_{Loss}}{\underbrace{||E_{x\sim p_{data}} f(x|y) - E_{z\sim p_{z}(z|y)}f(G(z))||^{2}_{2}}}  + \\ E_{x\sim p_{data}}[log(D(x|y))]
\end{aligned}
\end{equation}

where FM is the feature matching loss and the rest of the objective function is the binary cross entropy between true class label $y \in (0,1)$ and the predicted class probability. 
\subsection{Hyperparameter Settings} \label{parameterSettings1}

\begin{table}[b!]
\setlength{\tabcolsep}{3pt}
	\renewcommand{\arraystretch}{1.3}
	\begin{threeparttable}
		\begingroup\fontsize{10pt}{10pt}\selectfont
	\centering
	\caption{SDG-GAN hyperparameters settings}
	\label{hyperparametersovers}
\begin{tabular}{ll}
 \specialrule{0.2em}{0.2em}{0.2em} \textbf{Hyperparameters} & \textbf{Value}                                                                          \\ \hline
Learning Rate            & $1x10^{-4}$                                                                             \\
Optimiser                & $Adam $                                                                                 \\
Epochs                   & $100$                                                                                     \\
Batch Size               & $64$                                                                                      \\
Generator Layers         & \begin{tabular}[c]{@{}l@{}}$(Noise, 128), (128,64),$ \\ $(64, data size)$\end{tabular}      \\
Discriminator Layers     & \begin{tabular}[c]{@{}l@{}}$(data size, 128), (128,64), $\\ $(64,32), (32, 1)$\end{tabular} \\
Activation function      & $ReLU$                                                                                    \\
Noise Distribution       & $N(0,1)$                                                                                  \\
Noise                    & $50$ \\  \specialrule{0.2em}{0.2em}{0.2em}                                                                                   
\end{tabular}
\endgroup
\end{threeparttable}
\end{table}

Our proposed method has many hyperparameters that need to be tuned in order to achieve optimal performance. After experimenting with different set of hyperparameters, the hyperparameters below have been chosen after showing on producing the best results. We selected the settings presented in Table \ref{hyperparametersovers}. Future work could include optimising those hyperparameters for the oversampling task. The noise parameter distribution was set to be a Gaussian distribution  with size dimensions set to 50.  The dropout ratio was set to 0.2 on both discriminator's and generator's hidden layers. Batch size is 64 and number of epochs was set to 100. In terms of activation function rectified linear unit (ReLU) was used for the hidden layers where sigmoid for the output layer of discriminator and tanh for the ouptut layer of the generator. Adam optimiser was selected for the training \cite{2014adam}.
\section{Experimental Design} \label{methodValid}

To evaluate SDG-GAN as an oversampling method to tackle binary classification problems in imbalanced data, we compared the performance of the classification algorithms when combined with SDG-GAN and other state-of-the-art oversampling methods, e.g. SMOTE \cite{SMOTE}, ADASYN \cite{adasynCite} and B-SMOTE \cite{han2005borderline}, and other GAN-based oversampling architectures, e.g. cGAN. 

In Section \ref{benchData}, we introduce the publicly available datasets used as part of the evaluation process. In Section \ref{sdgog}, we apply our method to the real-world gambling dataset provided by our industrial partners, examining money laundering risk in online gambling.The following hypotheses need to be met to describe our method as successful:
\begin{itemize}
\item $H_{1}$: The use of SDG-GAN to augment imbalanced datasets will improve the algorithmic performance in baseline experiments on the benchmark imbalanced datasets.
\item $H_{2}$: The use of SDG-GAN will improve the algorithmic performance of classification algorithms in the real-world gambling dataset. 
\end{itemize}

 $H_{1}$ and $H_{2}$ are tested by combining the original and synthetic datasets with the four classification algorithms, i.e. LR, RF, XGBoost and MLP, in Section \ref{Results_oversa}.

\subsection{Benchmark Datasets}\label{benchData}

We evaluated our method on the different benchmark imbalanced datasets presented in Table \ref{RDatasets}. The IR was defined as the imbalance ratio between the minority and majority classes. We used data from different sectors to examine the range of applications for our method. We selected the Credit Card Fraud Dataset from Kaggle \cite{dal2017credit} and the Pima Diabetes \cite{diabetes} and Breast Cancer Wisconsin (Diagnostic) Datasets from the UCI Machine Learning Repository \cite{cancer}, an online resource containing several datasets for machine learning purposes. 

The rationale behind using the benchmark datasets was so that the results of this study could be easily compared to similar studies carried out previously and in the future. Moreover, it was decided that all datasets should describe a binary classification problem and contain numeric features to be in the same format as our gambling data. 

The Credit Card Fraud Dataset contains transactions made by credit cards in September 2013 by European cardholders. It presents transactions that occurred over two days, with 492 frauds out of the 2,492 transactions. The Wisconsin Breast Cancer Dataset includes features computed from a digitised image of a fine needle aspirate of a breast mass. The features describe characteristics of the cell nuclei present in the image. The purpose of the dataset is to classify a diagnosis as positive or negative. The PIMA Diabetes Dataset is originally from the National Institute of Diabetes and Digestive and Kidney Diseases. Its objective is to diagnostically predict whether a patient has diabetes based on certain diagnostic measurements included in the dataset. Several constraints were placed on the selection of these instances from a larger database; in particular, all patients are females at least 21 years old of Pima Indian heritage.


\begin{table}[t!]

	\centering
	\setlength{\tabcolsep}{3pt}
	\renewcommand{\arraystretch}{1.3}
	\begin{threeparttable}
	\centering
	\begingroup\fontsize{9pt}{9pt}\selectfont
	\caption{UCI datasets. There are three different sectors (B = business, L= life sciences). Number of features, number of instances, imbalance ratio}
	\label{RDatasets}
	\begin{tabular}{cccccc}
		 \specialrule{0.2em}{0.2em}{0.2em}
		\textbf{ID} & \textbf{Data Set} & \textbf{Sector} & \textbf{\#Features} & \textbf{\#Instances} & \textbf{IR} \\ \hline
		1           & Credit Card Fraud & B               & 30                  & 2,492                 & 1:4.07        \\
		2           & PIMA Diabetes     & L               & 8                   & 768                  & 1:1.87        \\
		3           & Breast Cancer     & L               & 30                  & 569                  & 1:1.68     \\ 
	      \specialrule{0.2em}{0.2em}{0.2em}
	\end{tabular}
		\endgroup
\end{threeparttable}
\end{table}

\begin{table}[t]
	\centering
		\setlength{\tabcolsep}{3pt}
	\renewcommand{\arraystretch}{1.3}
	\begin{threeparttable}
		\centering
		\begingroup\fontsize{9pt}{9pt}\selectfont
	\caption{Real-world Gambling Dataset}
	\label{Gambling D}
	\begin{tabular}{cccccl}
		 \specialrule{0.2em}{0.2em}{0.2em}
		\textbf{ID} & \textbf{Data Set} & \textbf{Sector} & \textbf{\#Features} & \textbf{\#Instances} & \textbf{IR} \\ \hline
		1           & Gambling Fraud  & B               & 31                 &  4,700                 & 1:2.97      \\
		 \specialrule{0.2em}{0.2em}{0.2em}
	\end{tabular}
		\endgroup
\end{threeparttable}
\end{table}

Note that before these datasets were used, their attribute values were scaled to be in interval $[0,1]$ by the min-max method to make the range of all attributes the same, preventing any one of them from dominating the others due to its scale. This reduced the range of values that the generator had to produce as well. With regard to implementation, all standard oversampling method tests were implemented using the `$imblearn.over\_sampling$' module in Python.  We used the default hyperparameter settings for SMOTE and its variants, i.e. $kneighbours = 5$.
For cGAN, we primarily used the hyperparameter settings that we set for SDG-GAN method, as seen in Table \ref{hyperparametersovers}.

\section{Results} \label{Results_oversa} 

For each dataset, we present the classification results observed after 10 runs for each oversampling technique and classification algorithm. The results in this section represent the average scores during those 10 runs. Similar to the process of \cite{tanaka2019data}, we split the data into testing and training sets. The training set included 80\% of the total population of the samples of each class and the testing set the other 20\% of the data. The data were shuffled to ensure reliable distribution in the sets. 

In the SDG-GAN, given an imbalanced training dataset, we first calculated the imbalance difference between the classes in the dataset. Then, a set of noise vectors with a dimension of 50 was used as the input for the generator. We trained the network generator by optimising the generator using the loss equation (\ref{lossFeature}). Real and synthetic data were then used as input for the discriminator D to output a probability value for evaluating the authenticity of the input data. Finally, the simulation samples generated through SDG-GAN were bonded with the original samples to enhance and balance the training dataset, which was then fed into the machine learning model for training.

\subsection{Results of Benchmark Datasets}\label{resultsPBD}

Table \ref{ccresults}, Table \ref{results_breast_cancer} and Table \ref{diabetes_res} show the results observed for the  three imbalanced public datasets of Credit Card Fraud, Breast Cancer and Pima Diabetes. We compare the five oversampling techniques in combination with  four classification algorithms. The performance of each classification method is measured in terms of recall, precision and F1 score.

\begin{table*}[ht!]
	\centering
	\caption{ Credit Card detection results: recall, precision and F1 measure}
	\label{ccresults}
	\begin{subtable}{1\textwidth}
		\centering
		\setlength{\tabcolsep}{4pt}
		\renewcommand{\arraystretch}{1.3}
		\begin{threeparttable}
			\begingroup\fontsize{12pt}{12pt}\selectfont
\begin{tabular}{cccccccc}
				\specialrule{0.2em}{0.2em}{0.2em}
				\textbf{Algorithms} & \textbf{Metrics} & \textbf{W/O} & \textbf{SMOTE} & \textbf{ADASYN} & \textbf{B-SMOTE} & \textbf{cGAN}  & \textbf{SDG-GAN} \\ 
				\hline
				LR & \begin{tabular}[c]{@{}c@{}}Recall\\ Precision\\ F1\end{tabular} & \begin{tabular}[c]{@{}c@{}}0.8142\\ \textbf{1.0000}\\ 0.8975\end{tabular} & \begin{tabular}[c]{@{}c@{}}0.8571\\ 0.9545\\ 0.9032\end{tabular} & \begin{tabular}[c]{@{}c@{}}0.8667\\ 0.7865\\ 0.8246\end{tabular} & \begin{tabular}[c]{@{}c@{}}0.8495\\ 0.9320\\ 0.8888\end{tabular} & \begin{tabular}[c]{@{}c@{}}0.8144\\ 0.9875\\ 0.8926\end{tabular} & \begin{tabular}[c]{@{}c@{}}	0.8090\\ 0.9863\\ 0.8889\end{tabular} \\ \hline
				
				RF & \begin{tabular}[c]{@{}c@{}}Recall\\ Precision\\ F1\end{tabular} & \begin{tabular}[c]{@{}c@{}}0.8984\\ 0.9170\\ 0.9076\end{tabular} & \begin{tabular}[c]{@{}c@{}}0.8694\\ 0.9586\\ 0.9116\end{tabular} & \begin{tabular}[c]{@{}c@{}}0.9288\\ 	0.8309\\ 0.8771\end{tabular} & \begin{tabular}[c]{@{}c@{}}0.8894\\ 0.9106\\ 0.8999\end{tabular} & \begin{tabular}[c]{@{}c@{}}0.8453\\ 0.9647\\ 0.9010\end{tabular} & \begin{tabular}[c]{@{}c@{}}0.9208\\ 0.9055 \\ \textbf{0.9131}\end{tabular} \\ \hline

				XGB & \begin{tabular}[c]{@{}c@{}}Recall\\ Precision\\ F1\end{tabular} & \begin{tabular}[c]{@{}c@{}} 0.8973\\ 0.9112\\0.9042\end{tabular} &
				\begin{tabular}[c]{@{}c@{}}		0.9163\\ 		0.8959\\ 	0.9060\end{tabular} & \begin{tabular}[c]{@{}c@{}}0.9087\\ 0.8787\\ 0.8935\end{tabular} &

				\begin{tabular}[c]{@{}c@{}} 		0.8776\\ 		0.8600\\  		0.8687\end{tabular} &
				
				\begin{tabular}[c]{@{}c@{}}	0.8559\\ 0.9694\\ 0.9091\end{tabular} &

				\begin{tabular}[c]{@{}c@{}}		  0.9053 \\ 0.9122\\0.9087\end{tabular} \\ \hline
				
				MLP & \begin{tabular}[c]{@{}c@{}}Recall\\ Precision\\ F1\end{tabular} & \begin{tabular}[c]{@{}c@{}}0.8191
\\ 0.9390\\ 
0.8750 \end{tabular} & \begin{tabular}[c]{@{}c@{}}	0.8830\\ 0.8384\\ 0.8601
\end{tabular} & \begin{tabular}[c]{@{}c@{}}0.9087\\ 	0.8536\\ 0.8803\end{tabular} & \begin{tabular}[c]{@{}c@{}}0.8761\\ 0.9082\\ 0.8919\end{tabular} & \begin{tabular}[c]{@{}c@{}}0.8454\\ 0.9879\\ 0.9111\end{tabular}  & \begin{tabular}[c]{@{}c@{}}\textbf{0.9487}
\\ 0.8315\\ 0.8862\end{tabular} \\ 
				\specialrule{0.2em}{0.2em}{0.2em}

			\end{tabular}
			\endgroup
		\end{threeparttable}
	\end{subtable}
	
	\bigskip
	
	\caption{Breast Cancer detection results: recall, precision and F1 measure}
	\label{results_breast_cancer}
	\begin{subtable}{1\textwidth}
		\centering
		\setlength{\tabcolsep}{4pt}
		\renewcommand{\arraystretch}{1.3}
		\begin{threeparttable}

			\begingroup\fontsize{12pt}{12pt}\selectfont
			\begin{tabular}{ccccccccc}
				\specialrule{0.2em}{0.2em}{0.2em}
				\textbf{Algorithm} & \textbf{Metrics}                                                        & \textbf{W/O}                                                     & \textbf{SMOTE}                                                    & \textbf{ADASYN}                                                  & \textbf{B-SMOTE}                                                      & \textbf{cGAN}                                                                                                       & \textbf{SDG-GAN}                                                   \\ \hline
				LR                 & \begin{tabular}[c]{@{}c@{}}Recall \\ Precision\\ F1\end{tabular} & \begin{tabular}[c]{@{}c@{}}0.8095\\ 0.9189\\ 0.8608\end{tabular}  & \begin{tabular}[c]{@{}c@{}}0.8888\\ 0.9302\\  0.9091\end{tabular} & \begin{tabular}[c]{@{}c@{}} 0.9048\\ 0.8636\\ 0.8837\end{tabular} & \begin{tabular}[c]{@{}c@{}}0.9302\\ 0.8888\\ 0.9090\end{tabular}  & \begin{tabular}[c]{@{}c@{}}0.9067\\ 0.8863\\ 0.8966\end{tabular} &  \begin{tabular}[c]{@{}c@{}}0.8837\\ 0.9500\\ 0.9157\end{tabular}   \\ \hline
				RF                 & \begin{tabular}[c]{@{}c@{}}Recall \\ Precision\\ F1\end{tabular} & \begin{tabular}[c]{@{}c@{}}0.8604\\ 0.8809\\ 0.8706\end{tabular}  & \begin{tabular}[c]{@{}c@{}}0.9069\\ 0.8478\\ 0.8764\end{tabular}  & \begin{tabular}[c]{@{}c@{}}0.8666\\ 0.9069\\ 0.8863\end{tabular} & \begin{tabular}[c]{@{}c@{}}0.9069\\ 0.8667\\ 0.8863\end{tabular}  & \begin{tabular}[c]{@{}c@{}}0.8604\\ 0.9024\\ 0.8809\end{tabular} &  \begin{tabular}[c]{@{}c@{}}0.8809\\  0.8604 \\ 0.8706\end{tabular} \\ \hline
				
				XGB & \begin{tabular}[c]{@{}c@{}}Recall\\ Precision\\ F1\end{tabular} & \begin{tabular}[c]{@{}c@{}}	0.8524\\0.8802\\0.8637\end{tabular} &

				\begin{tabular}[c]{@{}c@{}}		0.9262\\ 		0.8282\\  		 0.8742	\end{tabular} & \begin{tabular}[c]{@{}c@{}}0.9143\\ 0.8426\\ 0.8754\end{tabular} &

				\begin{tabular}[c]{@{}c@{}}  0.9119 \\ 0.8567\\ 0.8821\end{tabular} &
				
				\begin{tabular}[c]{@{}c@{}}0.9524
					
					\\ 0.8889\\ \textbf{0.9195}\end{tabular} &
				
				\begin{tabular}[c]{@{}c@{}}		 0.8571\\ 0.9767\\ 0.9130\end{tabular} \\ \hline

				MLP                & \begin{tabular}[c]{@{}c@{}}Recall \\ Precision\\ F1\end{tabular} & \begin{tabular}[c]{@{}c@{}}0.8604\\ \textbf{0.9487}\\ 0.9024\end{tabular}  & \begin{tabular}[c]{@{}c@{}}0.9069\\ 0.8863\\ 0.8965\end{tabular}  & \begin{tabular}[c]{@{}c@{}}\textbf{0.9381}\\ 0.7940\\ 0.8578\end{tabular} & \begin{tabular}[c]{@{}c@{}}0.9302\\ 0.8888\\ 0.9090\end{tabular}  & \begin{tabular}[c]{@{}c@{}}0.9069\\ 0.8863\\ 0.8965\end{tabular}   & \begin{tabular}[c]{@{}c@{}}0.8604\\ 0.9737\\ 0.9137\end{tabular}   \\ \specialrule{0.2em}{0.2em}{0.2em}
			\end{tabular}
			
			\endgroup
		\end{threeparttable}
	\end{subtable}

\end{table*}

For the Credit Card Fraud Dataset in Table \ref{ccresults}, the highest F1 score was achieved when SDG-GAN was combined with RF for a score of 91.31\%. In Table \ref{results_breast_cancer} for the Breast Cancer Dataset, cGAN combined with XGBoost outperformed the rest of the methods with F1 score of 91.95\%.  Similarly with the Credit Card, in the Pima Diabetes Dataset, SDG-GAN in combination with RF produced the best results with an F1 score of 70.80\% as Table \ref{diabetes_res} indicates. This was a significant improvement of $\approx$ 5\% compared to when no oversampling was used and an improvement of $\approx$ 2\% than the second-best combination between MLP and ADASYN. 
Another observation that could be drawn from the results was that when the standard oversampling techniques were used i.e. SMOTE, ADASYN, there was a drastic improvement in the classification of the minority class with better overall recall compared to precision (in the majority of cases). However, simultaneously, there was a huge drop in the classification accuracy of the majority class. This was supported by the increase of the recall score in the Credit Card Fraud Dataset prior to the use of any oversampling method; on average, the recall was 85\% and the precision 94\%. When SMOTE was used, the recall score increased significantly, while the precision decreased. However, this was not the case when SDG-GAN was used for oversampling, whereby we saw a more robust improvement in the classification metrics, as Table \ref{ccresults} and Table \ref{diabetes_res} show.

\begin{table*}[t!]
	\centering
	\setlength{\tabcolsep}{4pt}
	\renewcommand{\arraystretch}{1.3}
	\begin{threeparttable}
	\caption{Pima Diabetes Dataset results: recall, precision and F1 measure}
	\label{diabetes_res}
		\begingroup\fontsize{12pt}{12pt}\selectfont
	\begin{tabular}{cccccccc}
       \specialrule{0.2em}{0.2em}{0.2em}
		\textbf{Algorithm} & \textbf{Metrics}                                                        & \textbf{W/O}                                                      & \textbf{SMOTE}                                                     & \textbf{ADASYN}                                                   & \textbf{B-SMOTE}                                                      & \textbf{cGAN}                                                                                                         & \textbf{SDG-GAN}                                                 \\ \hline
		LR                 & \begin{tabular}[c]{@{}c@{}}Recall \\ Precision\\ F1\end{tabular} & \begin{tabular}[c]{@{}c@{}}0.6181\\ \textbf{0.6939}\\ 0.6538\end{tabular}   & \begin{tabular}[c]{@{}c@{}}0.7455\\ 0.5694\\ 0.6456\end{tabular}   & \begin{tabular}[c]{@{}c@{}}0.7272\\ 0.5479\\ 0.6250\end{tabular}  & \begin{tabular}[c]{@{}c@{}}0.7091\\ 0.5652\\ 0.6290\end{tabular}  & \begin{tabular}[c]{@{}c@{}}0.6727\\  0.6851\\ 0.6788\end{tabular}   & \begin{tabular}[c]{@{}c@{}}	0.6727\\ 0.6379\\ 0.6549
\end{tabular} \\\hline
		
		RF                 & \begin{tabular}[c]{@{}c@{}}Recall \\ Precision\\ F1\end{tabular} & \begin{tabular}[c]{@{}c@{}}0.6727\\ 0.6379\\ 0.6548\end{tabular}   & \begin{tabular}[c]{@{}c@{}}0.7636\\ 0.6086\\ 0.6774\end{tabular}   & \begin{tabular}[c]{@{}c@{}}0.7454\\ 0.5775\\ 0.6507\end{tabular}  & \begin{tabular}[c]{@{}c@{}}0.6727\\ 0.6271\\ 0.6491\end{tabular}  & \begin{tabular}[c]{@{}c@{}}0.6545\\ 0.6101\\ 0.6315\end{tabular}  &  \begin{tabular}[c]{@{}c@{}}0.7272\\ 0.6897\\ \textbf{0.7080}\end{tabular} \\\hline
	
		XGB & \begin{tabular}[c]{@{}c@{}}Recall\\ Precision\\ F1\end{tabular} & \begin{tabular}[c]{@{}c@{}}0.6727
			\\ 0.5781
			\\0.6218
		\end{tabular} &
		
		\begin{tabular}[c]{@{}c@{}}		0.7091\\ 		0.6094
			\\ 0.6555
		\end{tabular} & \begin{tabular}[c]{@{}c@{}}0.7636\\ 0.5753\\ 0.6562
		\end{tabular} &
		
		\begin{tabular}[c]{@{}c@{}} 		0.7455\\ 				0.5775\\  		0.6508\end{tabular} &
		
		\begin{tabular}[c]{@{}c@{}}0.6727
			\\ 0.6066
			\\ 0.6379\end{tabular} &
		
	 \begin{tabular}[c]{@{}c@{}}		 		0.6545\\ 		0.5902\\ 		0.6207\end{tabular} \\ \hline

		MLP                & \begin{tabular}[c]{@{}c@{}}Recall \\ Precision\\ F1 \end{tabular} & \begin{tabular}[c]{@{}c@{}}0.6182\\  0.6938\\  0.6538\end{tabular} & \begin{tabular}[c]{@{}c@{}}0.7818\\ 0.6142\\ 0.6880\end{tabular}   & \begin{tabular}[c]{@{}c@{}}\textbf{0.7818}\\ 0.5890\\ 0.6718\end{tabular}  & \begin{tabular}[c]{@{}c@{}}0.7091\\ 0.6094\\ 0.6555\end{tabular}  & \begin{tabular}[c]{@{}c@{}}0.6727\\  0.6491\\ 0.6607\end{tabular} & \begin{tabular}[c]{@{}c@{}}0.6727\\ 0.6852\\ 0.6788\end{tabular} \\ \specialrule{0.2em}{0.2em}{0.2em}
	\end{tabular}
	\endgroup
\end{threeparttable}
\end{table*}

The mean rankings of the F1 score per classifier across all datasets are presented in Table \ref{RankResultsFmeasure}. No one oversampling technique performed best across all classification methods and datasets. However, the SDG-GAN method performed consistently well and managed to achieve the highest overall mean rank score (2.6). 

Among the oversampling methods, SMOTE produced the second-best results, outperforming ADASYN and B-SMOTE. This indicated that the more recent variations of SMOTE do not necessarily outperform their predecessor, mirroring previous findings in the credit scoring literature \cite{engelmann2020conditional}.  Considering the mean ranking results from Table \ref{RankResultsFmeasure}, we could address the second hypothesis, $H_{1}$, stating that the use of SDG-GAN improves the classification performance in experiments on benchmark imbalanced datasets.

\begin{table}[ht!]
	\centering
	\setlength{\tabcolsep}{4pt}
	\renewcommand{\arraystretch}{1.3}
	\begin{threeparttable}
	\centering
	\caption{Summary Rank Results For F1 score}
	\label{RankResultsFmeasure}
	\begingroup\fontsize{12pt}{12pt}\selectfont
	\begin{tabular}{lllllll}
       \specialrule{0.2em}{0.2em}{0.2em}
		& \multicolumn{1}{c}{\textbf{Overall}} &  & \multicolumn{4}{c}{\textbf{Classifier}} \\ \cline{2-2} \cline{4-7} 
		Method & Mean Rank &  & \begin{tabular}[c]{@{}l@{}} LR\end{tabular} & \begin{tabular}[c]{@{}l@{}}RF\end{tabular} & XGB & \begin{tabular}[c]{@{}l@{}}MLP\end{tabular} \\ \hline
		SDG-GAN & \textbf{2.6} &  & 2.7 & 2.3 & 3.3 & 2.0 \\
		W/O & 4.3 &  & 3.3 & 4.0 & 5.0 & 4.7 \\
		SMOTE & 2.9 &  & 2.0 & 2.7 & 3.3 & 3.7 \\
		B-SMOTE & 3.6 &  & 4.0 & 3.7 & 3.7 & 3.0 \\
		ADASYN & 4.3 &  & 5.3 & 4.0 & 3.7 & 4.3 \\
		cGAN & 3.1 &  & 2.7& 4.3 & 2.0 & 4.0 \\        \specialrule{0.2em}{0.2em}{0.2em}
	\end{tabular}
	\endgroup
\end{threeparttable}
\end{table}

\section{SDG-GAN in Online Gambling}\label{sdgog}

After the success of our proposed method on the benchmark datasets, we applied our SDG-GAN technique for generating synthetic players’ data to tackle the imbalanced class in the real world fraud detection gambling dataset. As mentioned in Table \ref{Gambling D} we have 4,700 instances in the dataset from which 1,200 are described as high risk for money laundering. We compare our results with the existing system that our partners have in place with overall F1 score 84.7\%. 

We used SDG-GAN as part of the supervised learning framework for oversampling the minority class. Similar to the benchmark dataset case experiments, we evaluated the effectiveness of our approach for practical applications against the standard oversampling techniques and a GAN-based approach introduced in this paper. Table \ref{GAMBLING_DATA_set} presents the classification performance results for the gambling fraud dataset.

\begin{table*}[t!]
	
	\centering
	\setlength{\tabcolsep}{4pt}
	\renewcommand{\arraystretch}{1.3}
	\begin{threeparttable}
	\centering
	\caption{Gambling Dataset results}
	\label{GAMBLING_DATA_set}
	\begingroup\fontsize{12pt}{12pt}\selectfont
	\begin{tabular}{cccccccc}
		 \specialrule{0.2em}{0.2em}{0.2em}
		Algorithm & Metrics                                                                & W/O                                                              & SMOTE                                                            & ADASYN                                                           & B-SMOTE                                                          & cGAN                                                             & SDG-GAN                                                          \\ \hline
		LR        & \begin{tabular}[c]{@{}c@{}}Recall\\ Precision\\ F1\end{tabular} & \begin{tabular}[c]{@{}c@{}}0.6842\\ 0.8942\\ 0.7752\end{tabular} & \begin{tabular}[c]{@{}c@{}}0.8907\\ 0.8209\\ 0.8544\end{tabular} & \begin{tabular}[c]{@{}c@{}}0.8745\\ 0.8120\\ 0.8421\end{tabular} & \begin{tabular}[c]{@{}c@{}}0.9109\\ 0.7840\\ 0.8427\end{tabular} & \begin{tabular}[c]{@{}c@{}}0.6541\\ 0.8788\\ 0.7500\end{tabular} & \begin{tabular}[c]{@{}c@{}}0.7206\\ 0.8900\\ 0.7964\end{tabular} \\ \hline
		RF        & \begin{tabular}[c]{@{}c@{}}Recall\\ Precision\\ F1\end{tabular} & \begin{tabular}[c]{@{}c@{}}0.9245\\ 0.8546\\ 0.8881\end{tabular} & \begin{tabular}[c]{@{}c@{}}0.9338\\ 0.8389\\ 0.8838\end{tabular} & \begin{tabular}[c]{@{}c@{}}0.9249\\ 0.8328\\ 0.8764\end{tabular} & \begin{tabular}[c]{@{}c@{}}0.9367\\ 0.8223\\ 0.8761\end{tabular} & \begin{tabular}[c]{@{}c@{}}0.9245\\ 0.8556\\ 0.8887\end{tabular} & \begin{tabular}[c]{@{}c@{}}0.9004\\\textbf{ 0.8943}\\ \textbf{0.8973}\end{tabular} \\ \hline
		XGB       & \begin{tabular}[c]{@{}c@{}}Recall\\ Precision\\ F1\end{tabular} & \begin{tabular}[c]{@{}c@{}}0.9195\\ 0.8645\\ 0.8912\end{tabular} & \begin{tabular}[c]{@{}c@{}}0.9449\\ 0.8479\\ 0.8938\end{tabular} & \begin{tabular}[c]{@{}c@{}}0.9492\\ 0.8327\\ 0.8871\end{tabular} & \begin{tabular}[c]{@{}c@{}}0.9576\\ 0.8278\\ 0.8880\end{tabular} & \begin{tabular}[c]{@{}c@{}}0.8923\\ 0.8722\\ 0.8821\end{tabular} & \begin{tabular}[c]{@{}c@{}}0.9322\\ 0.8627\\ 0.8961\end{tabular} \\ \hline
		MLP       & \begin{tabular}[c]{@{}c@{}}Recall\\ Precision\\ F1\end{tabular} & \begin{tabular}[c]{@{}c@{}}0.8189\\ 0.8805\\ 0.8486\end{tabular} & \begin{tabular}[c]{@{}c@{}}0.9671\\ 0.7655\\ 0.8545\end{tabular} & \begin{tabular}[c]{@{}c@{}}0.9588\\ 0.7767\\ 0.8582\end{tabular} & \begin{tabular}[c]{@{}c@{}}\textbf{0.9712}\\ 0.7540\\ 0.8489\end{tabular} & \begin{tabular}[c]{@{}c@{}}0.8213\\ 0.8816\\ 0.8807\end{tabular} & \begin{tabular}[c]{@{}c@{}}0.8601\\ 0.8636\\ 0.8619\end{tabular} \\		 \specialrule{0.2em}{0.2em}{0.2em}
	\end{tabular}
	\endgroup
	\end{threeparttable}
\end{table*}

Similar to the experiments on the Credit Card Fraud and Diabetes Datasets, the performance of SDG-GAN was superior, with the highest F1 measure and precision at 89.73\% and 89.43\%, respectively. As Table \ref{GAMBLING_DATA_set} shows, SDG-GAN combined with XGBoost and RF outperformed the other oversampling techniques. However, when combined with LR, we did not expect it to improve the classification performance. Overall, the SDG-GAN results showed it can effectively estimate even complex data distributions. Furthermore, the results from Table \ref{GAMBLING_DATA_set} supported the final hypothesis, $H_{2}$, stating that the use of SDG-GAN could improve the identification rate of risk of money laundering in online gambling. 

Comparing the new classification results with SDG-GAN and the rule-based system, there was a significant F1 score improvement of around 5\%. Overall, with our oversampling method, we managed to reduce the number of both false positives and false negatives compared to the other techniques and enhanced the ability of the classification algorithm to distinguish the AML and Normal groups’ classes.

\section{Conclusion}
In this paper, we introduced SDG-GAN, an architecture based on GANs for generating synthetic data. Our method was compared against popular oversampling techniques i.e. SMOTE, B-SMOTE and ADASYN as well as other adversarial network architecture that has been used for generating new data i.e. cGANs. We evaluated the ability of SDG-GAN to produce high-quality synthetic data by comparing the algorithmic performance of four machine learning classification algorithms when combined with our method on three public imbalanced datasets and a real-world gambling fraud dataset. We found that the SDG-GAN oversampling compared favourably to the other oversampling methods and achieved the highest overall rank, as Table \ref{RankResultsFmeasure} shows. Our method outperformed SMOTE, ADASYN, B-SMOTE and cGAN on three out of the four examined imbalanced datasets, with the best performance achieved when it was combined with RF in two out of the three experiments. 

In the real-world gambling dataset, the application of SDG-GAN helped improve the identification rate by improving the F1 score by 5\% compared to the rule-based system and around 0.4\% compared to the other oversampling techniques. 


\bibliographystyle{./IEEEtran}

\bibliography{./IEEEexample}

\vspace{12pt}
\vspace{12pt}

\end{document}